\documentclass[sigconf, authorversion,nonacm]{acmart}
\usepackage{bm}

\AtBeginDocument{%
  \providecommand\BibTeX{{%
    Bib\TeX}}}

\setcopyright{none}
\copyrightyear{}
\acmYear{}
\acmDOI{}

\acmConference[]{}{}{}
\acmPrice{}
\acmISBN{}

\makeatletter
\def\@copyrightspace{\relax}
\makeatother
\begin{document}

\title{Knowledge from Uncertainty in Evidential Deep Learning}

\author{Cai Davies}
\affiliation{
  \institution{Cardiff University}
  \country{UK}
}

\author{Marc Roig Vilamala}
\affiliation{
  \institution{Cardiff University}
  \country{UK}
}

\author{Alun D. Preece}
\affiliation{
  \institution{Cardiff University}
  \country{UK}
}

\author{Federico Cerutti}
\affiliation{
  \institution{University of Brescia}
  \country{Italy}
}

\author{Lance M. Kaplan}
\affiliation{
  \institution{DEVCOM Army Research Lab}
  \country{USA}
}

\author{Supriyo Chakraborty}
\affiliation{
  \institution{IBM Research}
  \country{USA}
}

\renewcommand{\shortauthors}{Davies et al.}

\begin{abstract}
    This work reveals an {\em evidential signal} that emerges from the uncertainty value in Evidential Deep Learning (EDL). EDL is one example of a class of uncertainty-aware deep learning approaches designed to provide confidence (or epistemic uncertainty) about the current test sample. In particular for computer vision and bidirectional encoder large language models, the `evidential signal' arising from the Dirichlet strength in EDL can, in some cases, discriminate between classes, which is particularly strong when using large language models. We hypothesise that the KL regularisation term causes EDL to couple aleatoric and epistemic uncertainty. In this paper, we empirically investigate the correlations between misclassification and evaluated uncertainty, and show that EDL's `evidential signal' is due to misclassification bias. We critically evaluate EDL with other Dirichlet-based approaches, namely Generative Evidential Neural Networks (EDL-GEN) and Prior Networks, and show theoretically and empirically the differences between these loss functions. We conclude that EDL's coupling of uncertainty arises from these differences due to the use (or lack) of out-of-distribution samples during training.

\end{abstract}

\ccsdesc[500]{Computing methodologies~Reasoning about belief and knowledge}
\ccsdesc[300]{Computing methodologies~Natural language processing}
\ccsdesc[300]{Computing methodologies~Computer vision}

\keywords{evidential deep learning, epistemic uncertainty, uncertainty estimation}


\maketitle

\section{Introduction}
This work investigates an \textit{evidential signal} that seems to emerge from the evidential deep learning (EDL) approach.  EDL is part of a class techniques to realise uncertainty-aware deep models that are able to faithfully represent decision confidence. 

When an artificial intelligence (AI) system assists a human operator with predictions, the human has to develop insights (\textit{i.e.} a mental model) of when to trust the AI system with its recommendations \cite{DBLP:conf/aaai/BansalNKWLH19}. If the human mistakenly relies upon the AI system in regions where it is likely to err, catastrophic failures may occur \cite{Kocielnik2019,bansal2019beyond}. To identify such regions where the AI system is likely to err, we need to distinguish between (at least) two different sources of uncertainty: \emph{aleatoric} (or \emph{aleatory}) and \emph{epistemic}  uncertainty \cite{HORA1996217,hullermeier_AleatoricEpistemicUncertainty_21}. Aleatoric uncertainty refers to the variability in the outcome of an experiment due to inherently random effects (e.g. flipping a fair coin). Epistemic uncertainty refers to the epistemic state of the agent using the model; hence its lack of knowledge that \--- in principle \--- can be reduced based on additional data samples. 

Estimating model (epistemic) uncertainty is crucial for trust \cite{tomsett2020rapid}. Approaches to this problem include Ensemble Distribution Distillation \cite{malinin2019ensemble}, Posterior Networks \cite{charpentier2020posterior}, Prior Networks \cite{prior_networks}, Evidential Deep Learning (EDL) \cite{sensoy2018evidential}, and Evidential Deep Learning with GANs (EDL-GEN) \cite{sensoy_UncertaintyAwareDeepClassifiers_20a}, all of which aim at separating aleatoric and epistemic uncertainty. We focus our attention on the last three mentioned approaches.
EDL (Section \ref{sec:edl}), in particular, transforms a classification problem into a regression one, with the goal of computing pseudo-counts of evidence in favour of different classes, which then can be mapped into a Dirichlet distribution. Given a classification problem, Prior Networks (Section \ref{sec:prior}), also, returns a Dirichlet distribution, which is computed considering an additional dataset of out-of-distribution (OOD) samples. Finally, EDL-GEN (Section \ref{sec:edlgen}) expands the idea of Prior Networks by automatically synthesising out-of-distribution samples.

The goal of this paper is to investigate a peculiarity in the Dirichlet strength output by a model trained using EDL which we call an `evidential signal', which we observe to be strongly coupled with the ground truth class, and in some cases carries sufficient information to discriminate between classes on this signal alone. This is seen to be particularly strong when using large language models for text classification, where the discriminatory power of the `evidential signal' matches the accuracy of the model's classifications. We hypothesise that this is caused by misclassification biases, which are intrinsic in the datasets, indicating that aleatoric and epistemic uncertainty are unintentionally closely coupled. We theoretically and experimentally explore the differences between the EDL, EDL-GEN and Prior Networks approaches to quantifying epistemic uncertainty, and show that these latter two single pass methods do not exhibit this same signal, indicating a better biasing of the Dirichlet strength to represent epistemic uncertainty. We conclude this is due to these methods being heavily dependent on using out-of-distribution samples during training, where EDL is dependent on a KL regularisation term that is high when a misclassification occurs.

\textbf{The main contributions of this paper are therefore threefold:} (1) a critical account of the differences between EDL's, EDL-GEN's, and Prior Networks' loss functions which highlights how the Dirichlet strength of the distribution outputted by EDL is heavily dependent on misclassification bias; (2) an empirical evaluation confirming such a hypothesis and revealing how the Dirichlet strength is coupled with the class, in some cases carrying enough information for class discrimination matching the accuracy of the model; and (3) a confirmation that such a bias is missing or very weak in Prior Networks and EDL-GEN training with out-of-distribution samples.

\section{Quantifying Aleatoric and Epistemic Uncertainties: Neural Approaches}
\label{section:methods}

In this section, we focus on three approaches to equip neural networks with the ability to express aleatory and epistemic uncertainty in classification problems. \textbf{Evidential Deep Learning} (EDL) \cite{sensoy2018evidential}, probably one of the simplest approaches, we summarise in Section \ref{sec:edl}. \textbf{Prior Networks} \cite{prior_networks}, an approach that makes use of an explicit out-of-distribution dataset, see Section \ref{sec:prior}. {Evidential Deep Learning} with GANs \cite{sensoy_UncertaintyAwareDeepClassifiers_20a}, which builds on the previous two and avoids the need for a pre-chosen out-of-distribution data set while relying on a generative network to synthesise noisy data samples, Section \ref{sec:edlgen}.

\subsection{Evidential Deep Learning}
\label{sec:edl}
Evidential Deep Learning (EDL) \cite{sensoy2018evidential} replaces the classical use of a softmax as the final layer for classification with an estimation of 
the parameters of a Dirichlet distribution over the probability mass function for the possible classes. During training, the model pseudo-counts \textit{evidence}, which is a measure of the amount of support collected from the data in favour of a sample being classified into a particular class.

 From this evidence, the belief masses ($\bm{b}_k$) and the uncertainty ($\bm{u}$) for each class can be calculated as follows. Let $e_{k}\geq0$ be the evidence derived for the $k^{th}$ class:
        $b_{k}=\frac{e_{k}}{S}$ and $u=\frac{K}{S}$,
%
where $K$ is the number of classes and $S=\sum_{i=1}^{K}(e_{i}+1)$, which is the sum of evidence over all classes, is referred to as the Dirichlet strength. We can define the parameters of the output Dirichlet distribution of sample $i$ as $\alpha_{i}=f(\mathbf{x}_{i}|\Theta)+1$ where $f(\mathbf{x}_{i}|\Theta)$ represents the evidence vector of sample $i$ given the model parameters. 

During training, the model can discover patterns in the data and generate evidence for specific class labels such that the overall loss is minimised. However, these features may also be present in counter-examples giving rise to misleading evidence. Reducing the magnitude of generated evidence may increase the overall loss, despite reducing the loss contributed by these counter-examples. To combat this, a regularisation term is included, which incorporates a Kullback-Leibler (KL) divergence term between a uniform Dirichlet distribution and $\tilde{\alpha}$, where $\tilde{\alpha}$ is the parameters of the output Dirichlet distribution $\alpha$ after removing the non-misleading evidence from $f(\mathbf{x}_{i}|\Theta)$, such that a correctly classified sample with no evidence for other classes will generate $\tilde{\alpha}$ as a uniform Dirichlet distribution.

To learn the parameters $\Theta$ of a neural network, EDL defines the loss function as
\begin{equation*}
    \begin{array}{r c l}
    \mathcal{L}(\Theta) & = &\sum_{i=1}^{N}\mathcal{L}_{i}(\Theta) + \\
    & + & \lambda_{t}\sum_{i=1}^{N}KL[D(\mathbf{\pi_{i}}|\tilde{\alpha}_{i})\;\|\;D(\mathbf{\pi_{i}}|\langle1,...,1\rangle)]
    \end{array}
\label{equation:loss_kl}
\end{equation*}
where 
$\lambda_{t}=\min(1.0,t/annealing\_step)\in[0,1]$ is the annealing coefficient, $t$ in the index of the current training epoch, and $annealing\_step$ is the epoch index at which $\lambda_{t}=1$.

Several options for $\mathcal{L}_{i}(\Theta)$ have been considered from \cite{sensoy2018evidential}, while most of the analysis in the original paper is performed using
\begin{equation*} 
    \mathcal{L}_{i}(\Theta)=\sum_{j=1}^{K}y_{ij}(\log(S_{i}) - \log(\alpha_{ij}))
\label{equation:edl_log_loss}
\end{equation*}
where $y_{i}$ represents the one-hot vector encoding of the ground-truth label for sample $i$. Our preliminary analysis with other options for $\mathcal{L}_{i}(\Theta)$ shows consistency of results. We make use of a PyTorch implementation of these loss functions.\footnote{\url{https://github.com/dougbrion/pytorch-classification-uncertainty} (on Jan 26th 2023, available under MIT license)}

\subsection{Prior Networks}
\label{sec:prior}
Prior Networks \cite{prior_networks} \footnote{\url{https://github.com/KaosEngineer/PriorNetworks} (on Jan 26th 2023} differs from EDL in that it uses an explicit out-of-distribution dataset alongside the in-distribution dataset the model is trained on. 
A Prior Network  \cite{prior_networks} for classification parameterises a distribution over a simplex, typically a Dirichlet distribution due to its tractable analytic properties. 
A Prior Network that parameterises a Dirichlet will be referred to as a \emph{Dirichlet Prior Network} (DPN). Similarly to an EDL, a DPN will generate the concentration parameters $\bm{\alpha}$ of the Dirichlet distribution.

DPNs for multi-class classification are trained to minimise the KL divergence between the model and a sharp Dirichlet distribution focused on the appropriate class for in-distribution data, and between the model and a flat Dirichlet distribution for out-of-distribution data, chosen in accordance with the \textit{principle of insufficient reason}.

DPNs' loss functions are thus of the form:
\begin{equation*}
\begin{array}{r c l}
\mathcal{L}(\bm{\theta}) & = & \mathbb{E}_{{p_{in}}(\bm{x})}[KL[D(\bm{\mu}|\bm{\hat \alpha})||p(\bm{\mu}|\bm{x};\bm{\theta})]] +\\
& & + \mathbb{E}_{{p_{out}}(\bm{x})}[KL[{ D}(\bm{\mu}|\bm{\tilde \alpha})||{p}(\bm{\mu}|\bm{x};\bm{\theta})]]
\end{array}
\label{eqn:dpnloss}
\end{equation*}
where ${p}(\bm{\mu}|\bm{x};\bm{\theta})$ is the output of the neural network model.

Therefore, it is necessary to define the in-distribution targets $\bm{\hat \alpha}$ and out-of-distribution targets $\bm{\tilde \alpha}$ with a flat (uniform) Dirichlet distribution. 
Concerning the in-distribution target, learning sparse delta functions is challenging as the error surface becomes poorly suited for optimisation. In \cite{prior_networks}, the authors re-parameterise their approach to work with the Dirichlet strength $S$ and $\mu_c = \frac{\alpha_c}{S}$, and they smooth the target means by redistributing a small amount of probability $\epsilon$ density across the various classes.

Finally, the true out-of-domain distribution is unknown, and samples are unavailable. A standard solution is to use a different, real dataset as a set of samples from the out-of-domain distribution.

\subsection{EDL-GEN}
\label{sec:edlgen}

To avoid the need to identify a dataset from an out-of-domain distribution, EDL-GEN \footnote{\url{https://muratsensoy.github.io/uncertainty.html} (on Jan 26th 2023} revisits the idea of Prior Networks by automatically generating out-of-distribution samples adding noise to a latent representation of the in-distribution samples derived using a variational autoencoder \cite{kingma_AutoEncodingVariationalBayes_14}.
Indeed, EDL-GEN \cite{sensoy_UncertaintyAwareDeepClassifiers_20a} builds upon implicit density models~\cite{mohamed2016learning} and noise-contrastive estimation~\cite{gutmann2012noise} to derive Dirichlet parameters for samples.
Let us consider a classification problem with $K$ classes and assume that $P_{in}$, $P_k$, and $P_{out}$ represent respectively the data distributions of the training set, class $k$, and out-of-distribution samples, i.e., the samples that do not belong to any of the $K$ classes.
A convenient way to describe the density of samples from a class $k$ is to describe it relative to the density of some other reference data.
By using the same reference data for all classes in the training set, we desire to get comparable quantities for their density estimations.
While in the noise-contrastive estimation approaches~\cite{hafner2018reliable}, reference is usually provided by noisy data, EDL-GEN generalises to out-of-distribution samples.

Using the dummy labels $y$:
\begin{equation*}
\label{eq:log_ratio}
\frac{P_k(\bm{x})}{P_{out}(\bm{x})} = \frac{p(\bm{x}|y = k)}{p(\bm{x}|y=out)} = \frac{p(y=k|\bm{x})}{p(y=out|\bm{x})} \left(\frac{1-\pi_k}{\pi_k} \right)
\end{equation*}
where $\pi_k$ is the marginal probability $p(y=k)$ and $(1-\pi_k)/\pi_k$ can be approximated as the ratio of sample size, i.e., $n_k/n_{out}$, which in \cite{sensoy_UncertaintyAwareDeepClassifiers_20a} is taken as one.

EDL-GEN approximates the log density ratio $\log\big({P_k(\bm{x})}/{P_{out}(\bm{x})}\big)$ as the logit output of a binary classifier~\cite{mohamed2016learning}, which is trained to discriminate between the samples from $P_k$ and $P_{out}$.
To train such a network, in \cite{sensoy_UncertaintyAwareDeepClassifiers_20a} the authors use the Bernoulli (logarithmic) loss.

Similarly to the other two proposals, EDL-GEN also considers a regularisation term 
\begin{equation}
\label{eq:reg_loss}
\mathcal{L}_2(\theta | \bm{x} ) = \beta KL [D(\bm{p}_{-k}| \bm{\alpha}_{-k}) \mid\mid D(\bm{p}_{-k} |\bm{1}) ],
\end{equation}
where $\bm{p}_{-k}$ refer to the vector of probabilities $p_j$ such that $j \neq k$.
This term helps achieve a classifier to be totally uncertain in its misclassification, except near decision boundaries.

Finally, one of the main contributions of EDL-GEN concerns the generation of out-of-distribution samples by using the latent space of a Variational Auto-Encoder \cite{kingma_AutoEncodingVariationalBayes_14} as a proxy for semantic similarity between samples in input space.
For each $\bm{x}_i$ in training set, EDL-GEN sample a latent point $\bm{z}$ from $q_\theta(\bm{z}\mid \bm{x}_i)$ and perturb it by $\bm{\epsilon} \sim q_{\gamma}(\bm{\epsilon}|\bm{z})$, which is implemented as a multivariate Gaussian distribution $\mathcal{N}(\bm{0},G(\bm{z}))$, where $G(\cdot)$ is a fully connected neural network with non-negative output that is trained via
\begin{equation}
\small
\label{eq:G}
\max_{G} ~~ \mathbb{E}_{{\substack{q_\theta(\bm{z} \mid \bm{x}_i), \\ q_{\gamma}(\bm{\epsilon}|\bm{z}), \\ p_\phi(\bm{\bar{x}}_i \mid \bm{z} + \bm{\epsilon})}}} \big[ \underbrace{ \log D'(\bm{z}+\bm{\epsilon})}_{(a)} + \underbrace{\log (1 - D(\bm{\bar{x}}_i)}_{(b)})  \big], \\
\end{equation}
where  $\bm{\bar{x}}_i \sim p_\phi(\bm{\bar{x}}_i \mid \bm{z} + \bm{\epsilon})$ is the decoded out-of-distribution sample from the perturbed sample $\bm{z} + \bm{\epsilon}$. The discriminators $D$ and $D'$ 
%
%
%
%
are binary classifiers with \textit{sigmoid} output 
that try to distinguish real samples from the generated ones. That is, given an input, a discriminator gives as an output the probability that the sample is from the training set distribution.
In Eq.~\ref{eq:G}, (a) forces the generated points to be similar to the real latent points through making them indistinguishable by $D'$ in the latent space of the VAE and (b) encourages the generated samples to be distinguishable by $D$ in the input space.
%


\section{Experimentation Methodology}
\label{sec:experimental-methodology}

\subsection{Hypotheses}
\label{sec:hypotheses}

\textbf{EDL is heavily dependent on misclassification bias (H1)} as the KL regularisation reduces all evidence near decision boundaries where training data classes overlap. This conflates epistemic (Dirichlet strength) and aleatoric uncertainty. For this, we explore the relationship between recall, which is a measure of misclassification, and the calculated uncertainty outputted by the model. We hypothesise a high correlation for EDL and weaker or no correlation for EDL-GEN and Prior Networks, which would indicate a coupling of aleatoric and epistemic uncertainty in the EDL loss function.

\textbf{The Dirichlet strength computed by EDL is coupled with class (H2)}. Given the coupling of the Dirichlet strength of EDL and misclassifications, and that aleatoric uncertainty is inherent in the dataset, we hypothesise that the Dirichlet strength will be closely coupled with the ground truth class, to the extent that in some cases where the misclassification bias is particularly strong, we can discriminate between classes based on the Dirichlet strength alone. To quantify this, we train a set of simple models on the Dirichlet strength and record the accuracy across a test set to measure the separability of the classes, which would indicate a coupling to the Dirichlet strength and differences in misclassifications between classes. Given our hypothesis of weak or no correlation with EDL-GEN and Prior Networks, we assume that this separability will be much weaker or non-existent using these approaches. 

Finally, \textbf{the evidential signal does not appear in approaches that train with out-of-distribution samples (H3)}, namely EDL-GEN and Prior Networks, which follows from the observation that the above two hypotheses do not apply to these methods, or that the observation is much weaker. These methods learn to reduce the Dirichlet strength (or increase the epistemic uncertainty)  over the out-of-distribution data. 

\subsection{Implementation Details}
\label{sec:implementation}
Models are implemented in PyTorch \cite{pytorch} using the PyTorch Lightning \footnote{\url{https://github.com/PyTorchLightning/pytorch-lightning} (on May 10th 2023, Apache-2.0 license)} framework. For the natural language tasks, the BERT \cite{devlin2019bert} architecture is used with bert-base-uncased pretrained weights, where the model is then fine-tuned to the dataset. For the computer vision tasks, an implementation of LeNet \cite{lecun1998gradient} and VGG16 \cite{vgg16} are used, where VGG16 was used as both fine-tuning using pretrained weights and full training with randomly initialised weights. 

AdamW \cite{adamw} was used as an optimiser. It is similar to the Adam optimiser \cite{adam_original} used in \cite{sensoy2018evidential} but shown to yield better training loss and improve generalisation. The learning rate usually performed well at $1\times10^{-3}$. The language models are fine-tuned for 30 epochs, while the full training of the computer vision models is 500 epochs.

\subsection{Datasets}

We use a selection of publicly available and widely used datasets in both the natural language and computer vision domains (Table~\ref{table:dataset}). The diversity of datasets includes variations between natural language and computer vision, as well as number of classes and classification types, e.g., sentiment, topic, gender etc.

\begin{table}[t]
\caption{Datasets used}
\label{table:dataset}
\vskip 0.15in
\begin{center}
\begin{small}
\begin{sc}
\begin{tabular}{lcccr}
\toprule
Dataset & Classes & Label \\
\midrule
Amazon \cite{ni2019justifying}   & 2 & sentiment \\
Blog \cite{schler2006effects} & 2 & gender\\
IMDB \cite{maas2011learning}    & 2 & sentiment \\
MNIST \cite{lecun1998gradient}    & 10 & digit        \\
Newsgroups \cite{Lang95}    & 20 & topic \\
\bottomrule
\end{tabular}
\end{sc}
\end{small}
\end{center}
\vskip -0.1in
\end{table}

\section{Experimental Validation of Hypotheses}
\label{section:validation}
\subsection{EDL is heavily dependent on misclassification bias}
\label{sec:misclassification}

\begin{figure}
    \centering
    \includegraphics[width=0.4\textwidth]{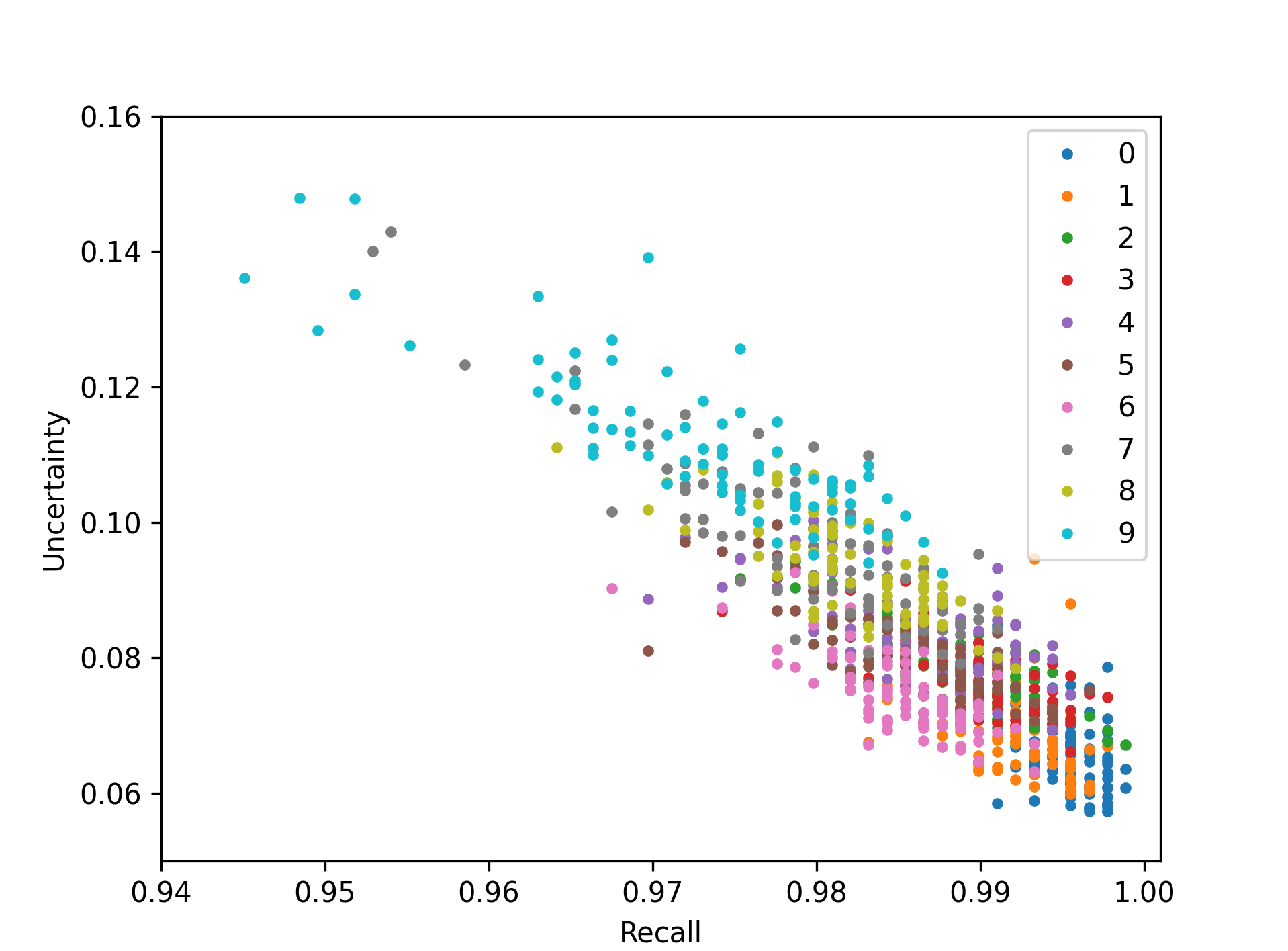}
    \caption{\textbf{Top:} Recall against evidential signal plot across models trained on varying annealing\_step values to highlight the correlation. \textbf{Bottom:} Pearson's and Spearman's correlation between recall and uncertainty.}
    \label{fig:mnist_recall}
\end{figure}

To investigate empirically hypothesis 1 (H1), we look to correlations between recall and the Dirichlet strength, or `evidential signal'. In Figure \ref{fig:mnist_recall} we show the results of $90$ LeNet models trained on the MNIST dataset, as defined in Section \ref{sec:implementation}, and using a random seed for training initalisations; on the x-axis we plot the recall, and y-axis we plot the uncertainty (or Dirichlet strength), and colour-code by ground truth class. There is a strong correlation between the recall and Dirichlet strength independent of the training seed, but also we notice a separation in the classes, in that digit $0$ tends to have high recall-low Dirichlet strength and in opposite to digit $9$, which indicates evidence towards H2. 

To quantify this correlation, we show the Pearson's and Spearman's rank correlation coefficients in Table \ref{table:correlation_table}. H1 hypothesises that this correlation will not be seen in EDL-GEN and Prior Networks, so we also show the results for these by running the experiment in the same manner for these two approaches. These results show EDL having a much higher correlation, close to a perfectly negative linear relationship for the Pearson's coefficient, where EDL-GEN and Prior Networks exhibit much weaker correlations. 

\begin{table}
\caption{Correlation between recall and evidential signal}
\centering
\begin{tabular}{l l l l l}
\toprule
\textbf{Approach} & \textbf{Pearson's} & \textbf{Spearman's}  \\
\midrule
EDL & -0.9975 & -0.8517 \\ 
EDL-GEN & -0.4899 & -0.4321 \\ 
Prior Networks & -0.4030 & -0.4320 \\ 
\bottomrule
\end{tabular}
\label{table:correlation_table}
\end{table}

\subsection{The Dirichlet Strength computed by EDL is coupled with class}
\label{sec:class_discrimination}
\begin{figure}
    \centering
    \includegraphics[width=0.4\textwidth]{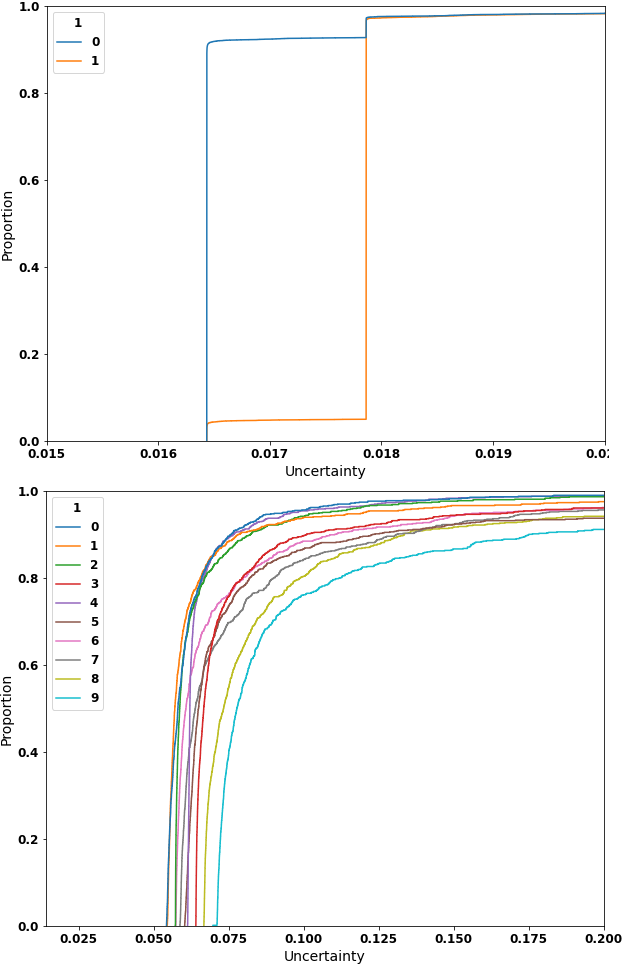}
    \caption{Cumulative evidential signal of samples split by the ground-truth label for IMDB (top) and MNIST (bottom).}
    \label{fig:imdb_mnist_single}
\end{figure}

As observed in Figure \ref{fig:mnist_recall}, across multiple runs, classes in MNIST appear to show a coupling with the Dirichlet strength and recall. To investigate this further, we first look to individual models, by creating a cumulative distribution function from the computed Dirichlet strength, or `evidential signal', of the samples in the test set and split by ground truth class. We show in Figure \ref{fig:imdb_mnist_single} the results of this for MNIST and the IMDB dataset, which is trained as described in Section \ref{sec:implementation}, to show that this coupling is not exclusive to MNIST or computer vision tasks. 


With IMDB, we see a very clear separation in the classes from the Dirichlet strength, where each appears to have `collapsed' into a single value; the crossover seen is due to samples being misclassified, which we can remove by separating by predicted class rather than ground truth. For MNIST, we see separation but of a much lower magnitude, with the Dirichlet strength being more distributed.

To quantify the class discrimination, and so the magnitude of the coupling between Dirichlet strength and class, we use three simple machine learning methods --- SVM, Decision Tree (DT) classifier \cite{Bishop_2006} and XGBoost \cite{Chen:2016:XST:2939672.2939785} --- to train on the `evidential signal' alone; we do this for each dataset from Table \ref{table:dataset} in accordance with Section \ref{sec:implementation} for 50 runs. From the test set of each dataset, we compute the Dirichlet strength and split this into a sub-train and sub-test set with an 80:20 ratio, which is used to train and test each of the simple ML methods keeping the original labels. A high accuracy would indicate strong separability; we show these results in Table \ref{table:uncertainty_separation}.

As in Section \ref{sec:misclassification}, the natural language tasks exhibit the clearest separability, with most of the datasets (with the exception of the multi-class Newsgroups) showing separability in the Dirichlet strength on par with the model's accuracy. From Figure \ref{fig:imdb_mnist_single} \textbf{bottom} we see the Dirichlet strength being more distributed than the natural language tasks, which is reflected in the poor separability scores with a maximum of $0.38$; however this is above the random-guessing threshold. 

\begin{table}
\caption{Separability of evidential signal across simple models by test accuracy}
\centering
\begin{tabular}{l l l l l}
\toprule
\textbf{Dataset} & \textbf{SVM}    & \textbf{DT} & \textbf{XGBoost} & \textbf{Orig} \\
\midrule
IMDB             & \textbf{0.9426} & 0.9396 & 0.9382 & 0.9400 \\ 
Blog             & \textbf{0.7746} & 0.7684 & 0.7670 & 0.7702 \\ 
Amazon     & \textbf{0.9274} & 0.9230 & 0.9264 & 0.9262 \\ 
Newsgroups       & 0.2734 & 0.7326 & \textbf{0.7624} & 0.8007  \\ 
MNIST            & 0.1424 & 0.2657 & \textbf{0.3789} & 0.9813 \\  
\bottomrule
\end{tabular}
\label{table:uncertainty_separation}
\end{table} 

\subsection{Evidential signal does not appear in OOD approaches}
\label{section:ood_approaches}

In Section \ref{sec:misclassification} we show the weaker correlation between the recall and Dirichlet strength for the out-of-distribution methods. To explore whether the `evidential signal' observed in EDL is seen in these other approaches ({H3), we run the same experiment as in Section \ref{sec:class_discrimination} with implementations of EDL-GEN \cite{sensoy_UncertaintyAwareDeepClassifiers_20a} and Prior Networks \cite{prior_networks}. Given that Prior Networks uses an explicit out-of-distribution dataset (see Section \ref{sec:prior}), we choose pairs from our existing datasets for NLP classifications, for example, we take IMDB as the in-distribution and Amazon Books as the out-of-distribution. For MNIST, we use EMNIST handwritten letters as the out-of-distribution dataset. Figures \ref{fig:prior_networks_mnist} and \ref{fig:edl_gen_mnist} show these results for MNIST using EDL-GEN and Prior Networks, respectively. Due to the method for generating out-of-distribution samples seen in EDL-GEN, the application of this method to the NLP datasets is not trivial, and so these experiments were omitted.

\begin{figure}
    \centering
    \includegraphics[width=0.4\textwidth]{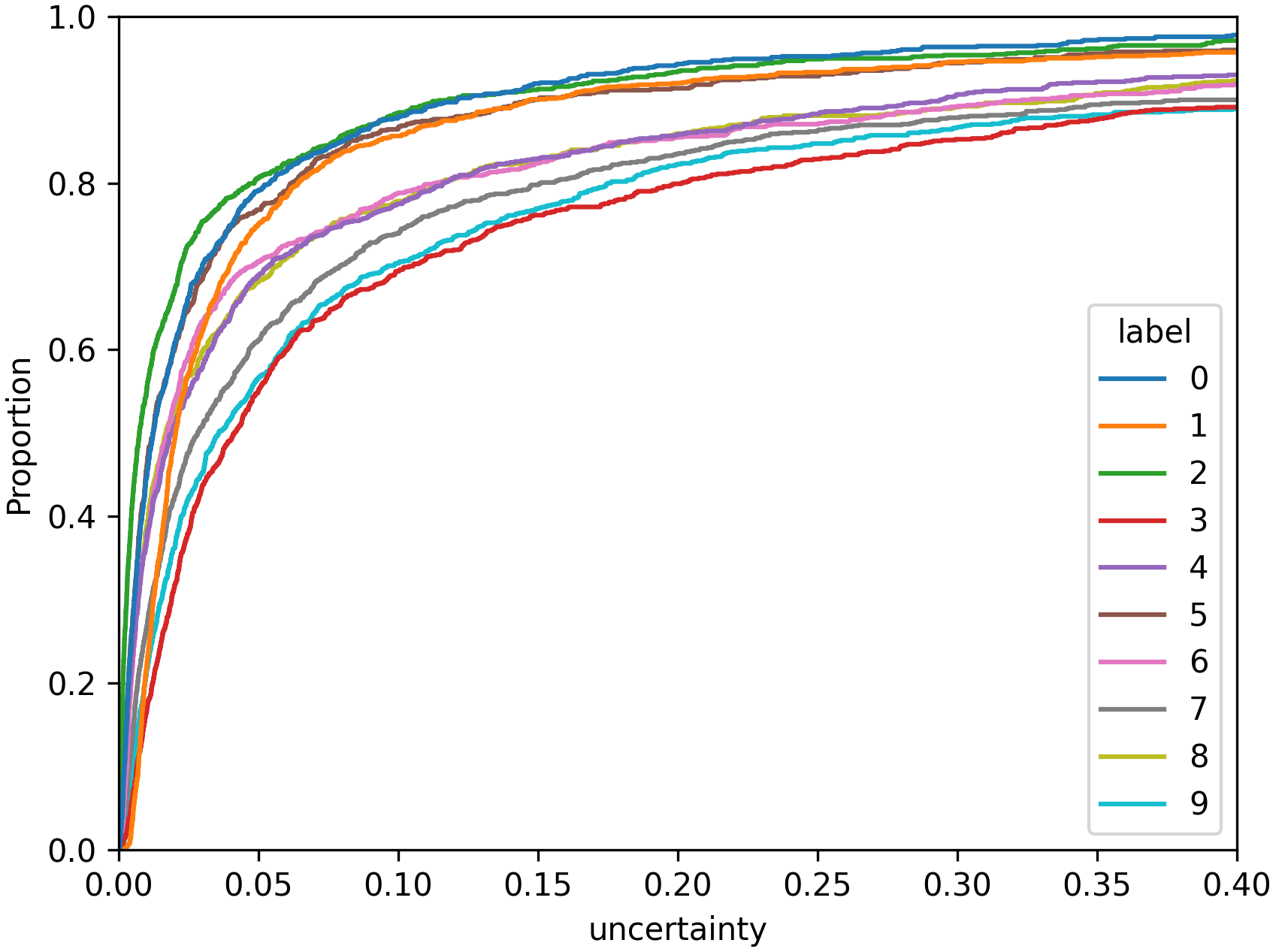}
    \caption{EDL-GEN implementation with MNIST and computed uncertainty plotted as a cumulative distribution function.}
    \label{fig:edl_gen_mnist}
\end{figure}  

\begin{figure}
    \centering
    \includegraphics[width=0.4\textwidth]{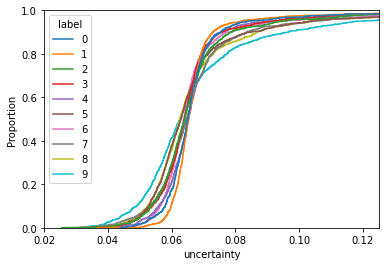}
    \caption{Prior Networks implementation with MNIST and computed uncertainty plotted as a cumulative distribution function.}
    \label{fig:prior_networks_mnist}
\end{figure}

Visually, the separability of classes, or class discrimination from the Dirichlet strength, appears non-existent. To quantify this, as before, we use the same set of simple machine learning methods trained on the Dirichlet strength and take the mean accuracy of the sub-test set across $50$ runs to indicate separability, which is shown in Table \ref{table:uncertainty_separation_all}. To simplify, we show the best mean accuracy of the three methods. As in Table \ref{table:uncertainty_separation}, EDL shows very good separation for the natural language tasks (IMDB), and poor but above random-guessing for MNIST. For EDL-GEN and Prior Networks, we see scores that are close to random-guessing, confirming the visual observation that the separability is weak or non-existent, which is in-line with H3 that this `evidential signal' does not appear in approaches that train using out-of-distribution samples.


\begin{table}
\caption{Best separability of evidential signal across Dirichlet approaches}
\centering
\begin{tabular}{l l l l l}
\toprule
\textbf{Approach} & \textbf{IMDB}    & \textbf{MNIST} \\
\midrule
EDL            & 0.9426 & 0.3789   \\ 
EDL-GEN        & N/A & 0.1465  \\ 
Prior Networks & 0.55 & 0.1515 \\
\bottomrule
\end{tabular}
\label{table:uncertainty_separation_all}
\end{table}

\section{Conclusion and Future Work}
\label{section:conclusion}
In this paper, we have critically explored differences between EDL, EDL-GEN and Prior Networks both theoretically (Section \ref{sec:hypotheses}) and empirically (Section \ref{section:validation}). We explored the relationship between recall and the Dirichlet strength of EDL, showing a strong correlation not seen in the other two approaches, which was further shown by the discriminatory power of the evidential signal in EDL only. Since these observations are due to misclassification bias, we have shown that the output of EDL is coupled with aleatoric uncertainty, which is inherent in the dataset. As such, the computed uncertainty differs to these other approaches, given the loss functions of EDL-GEN and Prior Networks heavily depends on out-of-distribution samples to calculate the Dirichlet strength. 

Future work will address the limitations of the present study. We will explore additional choices of loss function from \cite{sensoy2018evidential} (preliminary analysis is consistent with our observation in Section \ref{sec:class_discrimination}). We will expand the investigation of the relationship between the evidential signal and misclassification bias to determine the cause of the NLP models exhibiting much stronger discrimination even in the multi-class case (i.e., Newsgroups). Finally, we will look for the presence of the evidential signal in other approaches for estimating epistemic uncertainty beyond EDL and Prior Networks.

\bibliographystyle{ACM-Reference-Format}
\bibliography{bib.bib}

\end{document}